\documentclass[conference]{IEEEtran}
\pdfoutput=1
\usepackage{times}
\usepackage[utf8]{inputenc}
\usepackage[sort&compress,numbers]{natbib}
\usepackage{multicol}
\usepackage[ruled,vlined,linesnumbered]{algorithm2e}
\usepackage[noend]{algpseudocode}

\usepackage{amsfonts}
\usepackage{amssymb,amsmath}
\usepackage{graphicx}
\usepackage{subcaption}
\usepackage[font=footnotesize]{subcaption}
\usepackage[font=footnotesize]{caption}
\usepackage{color}
\usepackage{booktabs} %for table rulers
\usepackage{tabulary}
\usepackage{lipsum}

\usepackage{amsthm} % for new theorem
\usepackage{thmtools}
\usepackage{mathtools}

\usepackage{hyperref}
\usepackage[usenames,dvipsnames,svgnames,table]{xcolor}
\usepackage{colortbl}
\definecolor{lightgreen}{rgb}{0.8,1,0.8}
\newcommand{\cmark}{\cellcolor{lightgreen}}%

\hypersetup{bookmarksopen,bookmarksnumbered,
pdfpagemode=UseOutlines,
colorlinks=true,
linkcolor=blue,
anchorcolor=blue,
citecolor=blue,
filecolor=blue,
menucolor=blue,
urlcolor=blue
}

\graphicspath{{figs/}}

\newcommand{\explicitGraph}[0]{G}
\newcommand{\vertexSet}[0]{V}
\newcommand{\edgeSet}[0]{E}
\newcommand{\vertex}[0]{v}
\newcommand{\edge}[0]{e}
\newcommand{\start}[0]{v_s}
\newcommand{\goal}[0]{v_g}

\newcommand{\length}[0]{\ell}

\renewcommand{\path}[0]{\xi}

\newcommand{\pathSet}[0]{\Xi}

\newcommand{\world}[0]{\phi}

\newcommand{\evalEdges}[0]{E_{\mathrm{eval}}}

\newcommand{\edgesValid}[0]{E_{\mathrm{val}}}
\newcommand{\edgesInvalid}[0]{E_{\mathrm{inv}}}

\newcommand{\stateSpace}{\mathcal{S}}
\newcommand{\actionSpace}{\mathcal{A}}
\newcommand{\transFn}{\mathrm{T}}
\newcommand{\rewardFn}{\mathrm{R}}
\newcommand{\state}{s}
\newcommand{\action}{a}

\newcommand{\stateAbs}{\mathcal{G}}

\newcommand{\dataset}{\mathcal{D}}
\newcommand{\mixParam}{\beta}
\newcommand{\policy}[0]{\pi}
\newcommand{\policySpace}[0]{\Pi}

\newcommand{\policyLearn}[0]{\hat{\policy}}
\newcommand{\policyMix}[0]{\policy_\mathrm{mix}}
\newcommand{\policyOracle}[0]{\policy_\mathrm{OR}}

\newcommand{\policyRoll}[0]{\pi_{\mathrm{roll}}}

\DeclareMathOperator*{\argmax}{arg\,max}

\newcommand{\maxprob}[1]{\underset{#1}{\max}}
\newcommand{\minprob}[1]{\underset{#1}{\min}}

\newcommand{\argmaxprob}[1]{\underset{#1}{\argmax}}

\newcommand{\abs}[1]{\left|#1\right|}
\newcommand{\card}[1]{\left|#1\right|}

\DeclareMathOperator*{\suchthat}{\;\; \mbox{s.t.} \;\;}
\newcommand{\expect}[2]{\mathbb{E}_{#1}\left[#2\right]}

\newcommand{\real}[0]{\mathbb{R}}

\newcommand{\bbm}{\begin{bmatrix}}
\newcommand{\ebm}{\end{bmatrix}}

\newcommand{\pair}[2]{\left( #1, #2\right)}

\newcommand{\seq}[2]{\left(#1_{1}, #1_{2}, \ldots, #1_{#2}\right)}

\newcommand{\setst}[2]{\left\lbrace #1\;\middle|\;#2\right\rbrace}

\newcommand{\Ind}[0]{\mathbb{I}}

\newcommand{\daggerAlg}[0]{\textsc{DAgger}}
\newcommand{\supervisedAlg}[0]{\textsc{Supervised}}

\newcommand{\algName}[0]{\textsc{StrOLL}\xspace}
\newcommand{\algFullName}[0]{\textbf{S}earch \textbf{t}h\textbf{r}ough \textbf{O}racle \textbf{L}earning and \textbf{L}aziness \xspace}
\newcommand{\algHeuristic}[0]{\textsc{StrOLL-R}\xspace}

\newcommand{\bisect}[0]{\textsc{BiSECt}\xspace}
\newcommand{\direct}[0]{\textsc{DiRECt}\xspace}

\newcommand{\lazysp}[0]{\textsc{LazySP}\xspace}

\newcommand{\selector}[0]{\textsc{Selector}\xspace}

\newcommand{\selectorOracle}[0]{\textsc{Oracle}\xspace}

\newcommand{\selectorRandom}[0]{\textsc{Random}\xspace}
\newcommand{\selectorForward}[0]{\textsc{Forward}\xspace}
\newcommand{\selectorBackward}[0]{\textsc{Backward}\xspace}
\newcommand{\selectorAlternate}[0]{\textsc{Alternate}\xspace}
\newcommand{\selectorFailFast}[0]{\textsc{FailFast}\xspace}
\newcommand{\selectorPostFailFast}[0]{\textsc{PostFailFast}\xspace}

\newcommand{\featPost}[0]{\textsc{Posterior}\xspace}
\newcommand{\featPrior}[0]{\textsc{Prior}\xspace}
\newcommand{\featIndex}[0]{\textsc{Location}\xspace}
\newcommand{\featDeltaLength}[0]{$\Delta$-\textsc{Length}\xspace}
\newcommand{\featDeltaEval}[0]{$\Delta$-\textsc{Eval}\xspace}
\newcommand{\featPDL}[0]{P\featDeltaLength}

\newcommand{\selectorPDL}[0]{P\featDeltaLength\xspace}

\newcommand{\eref}[1]{(\ref{#1})}
\newcommand{\sref}[1]{Section~\ref{#1}}
\newcommand{\figref}[1]{Fig.~\ref{#1}}
\newcommand{\algoref}[1]{Algorithm~\ref{#1}}
\newcommand{\algolineref}[1]{Line~\ref{#1}}

\newtheorem{theorem}{Theorem}

\newtheorem{problem}{Problem}

\newtheoremstyle{hypstyle}
{3pt} % Space above
{3pt} % Space below
{\itshape} % Body font
{} % Indent amount
{\bfseries} % Theorem head font
{.} % Punctuation after theorem head
{.5em} % Space after theorem head
{} % Theorem head spec (can be left empty, meaning `normal')

\theoremstyle{hypstyle} 
\newtheorem{observation}{O}
\newtheorem{ques}{Q}

\newcommand{\xxnote}[3]{}
\ifx\hidenotes\undefined
  \renewcommand{\xxnote}[3]{\color{#2}{#1: #3}}
\fi

\newcommand{\fullFigGap}[0]{\vspace{-1.5\baselineskip}} %Use this command at the end of the caption of a full-page figureto eat up some unnecessary whitespace.

\begin{document}

% paper title
\title{Leveraging Experience in Lazy Search}

% You will get a Paper-ID when submitting a pdf file to the conference system
%\author{Mohak Bhardwaj, Sanjiban Choudhury, Byron Boots, Siddhartha Srinivasa}
%\author{\authorblockN{Michael Shell}
%\authorblockA{School of Electrical and\\Computer Engineering\\
%Georgia Institute of Technology\\
%Atlanta, Georgia 30332--0250\\
%Email: mshell@ece.gatech.edu}
%\and
%\authorblockN{Homer Simpson}
%\authorblockA{Twentieth Century Fox\\
%Springfield, USA\\
%Email: homer@thesimpsons.com}
%\and
%\authorblockN{James Kirk\\ and Montgomery Scott}
%\authorblockA{Starfleet Academy\\
%San Francisco, California 96678-2391\\
%Telephone: (800) 555--1212\\
%Fax: (888) 555--1212}}

% for over three affiliations, or if they all won't fit within the width
% of the page, use this alternative format:
% 
\author{\authorblockN{Mohak Bhardwaj \authorrefmark{1},
Sanjiban Choudhury \authorrefmark{2},
Byron Boots \authorrefmark{1} and 
Siddhartha Srinivasa \authorrefmark{2}}
\authorblockA{\authorrefmark{1}{Georgia Institute of Technology} \authorrefmark{2}{University of Washington}}}

%Email: \{mbhardwaj8, bboots \}@cc.gatech.edu
%Email: \{sanjibac\}@thesimpsons.com}
\maketitle
\vspace{-4mm}
% !TEX root = ../main.tex
\begin{abstract}
Lazy graph search algorithms are efficient at solving motion planning problems where edge evaluation is the computational bottleneck. 
These algorithms work by lazily computing the shortest potentially feasible path, evaluating edges along that path, and repeating until a feasible path is found. 
The order in which edges are selected is critical to minimizing the total number of edge evaluations: 
a good edge selector chooses edges that are not only likely to be invalid, but also eliminates future paths from consideration.
We wish to learn such a selector by leveraging prior experience. 
We formulate this problem as a Markov Decision Process (MDP) on the state of the search problem. 
While solving this large MDP is generally intractable, % and approximation techniques have poor empirical convergence.
 we show that %if the latent edge status are known, 
 we can compute oracular selectors that can solve the MDP during training.
With access to such oracles, we  use imitation learning to find effective policies. If new search problems are sufficiently similar to problems solved during training, the learned policy will choose a good edge evaluation ordering and solve the motion planning problem quickly. 
%The learning offers valuable insights on which features are relevant to make such selections. 
We evaluate our algorithms on a wide range of $2$D and $7$D problems and show that the learned selector outperforms baseline commonly used heuristics. 
\end{abstract}

\IEEEpeerreviewmaketitle

% !TEX root = ../main.tex

\section{Introduction}
\label{sec:introduction}

% P1: \textbf{What is the problem?} We focus on the problem of finding the shortest path on a graph while minimizing planning effort. For problems such as robotic motion planning, edge evaluation is the bottleneck. Recently, LazySP has been shown to solve such problems efficiently - at every iteration, the current shortest path is selected and only edges belonging to it are evaluated. Either the edges are valid and path is verified to be valid, else the path is eliminated and the process continues. Our goal is to learn a policy that chooses which edge to evaluate such that the shortest feasible path is found after a minimum number of evaluations. (Fig 1)

\begin{figure}[!t]
\centering
\includegraphics[width=\columnwidth]{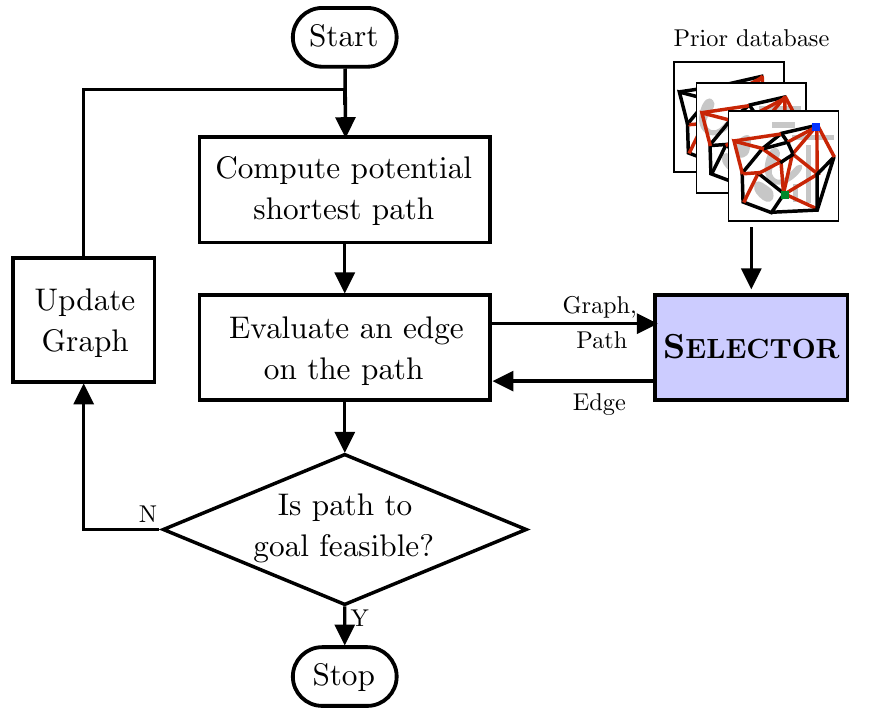}
\caption{The \lazysp~\citep{dellin2016unifying} framework. \lazysp iteratively computes the shortest path, queries a \selector for an edge on the path, evaluates it and updates the graph until a feasible path is found. The number of edges evaluated depends on the choice of \selector. We propose to train a \selector from prior data. \fullFigGap}
\label{fig:intro}
\end{figure}

In this paper, we explore algorithms that leverage past experience to find the shortest path on a graph while minimizing planning time.  
We focus on the domain of robot motion planning where the planning time is dominated by \emph{edge evaluation}~\citep{hauser2015lazy}.
Here the goal is to check the minimal number of edges, invalidating potential shortest paths along the way, until we discover the shortest feasible path -- this is the central tenet of lazy search~\citep{dellin2016unifying,bohlin2000path}. 
We propose to \emph{learn within this framework} which edges to evaluate (\figref{fig:intro}).

How should we leverage experience? Consider the ``Piano Mover's Problem''~\citep{schwartz1983piano} where the goal is to plan a path for a piano from one room in a house to another. Collision checking all possible motions of the piano can be quite time-consuming. Instead, what can we infer if we were given a database of houses and edge evaluations results?
\begin{enumerate}
	\item \emph{Check doors first} - these edges serve as bottlenecks for many paths which can be eliminated early if invalid.
	\item \emph{Prioritize narrow doors} - these edges are more likely to be invalid and can save checking other edges.
	\item \emph{Similar doors, similar outcomes} - these edges are correlated, checking one reveals information about others.
\end{enumerate}
Intuitively, we need to consider all past discoveries about edges to make a decision. While this has been explored in the Bayesian setting~\citep{choudhury2017active,choudhury2018bayesian}, we show that more generally the problem can be mapped to a Markov Decision Process (MDP).
However, the size of the MDP grows exponentially with the size of the graph. Even if we were to use approximate dynamic programming, we still need to explore an inordinate number of states to learn a reasonable policy.  

\begin{figure*}[!t]
    \centering
    \includegraphics[width=\textwidth]{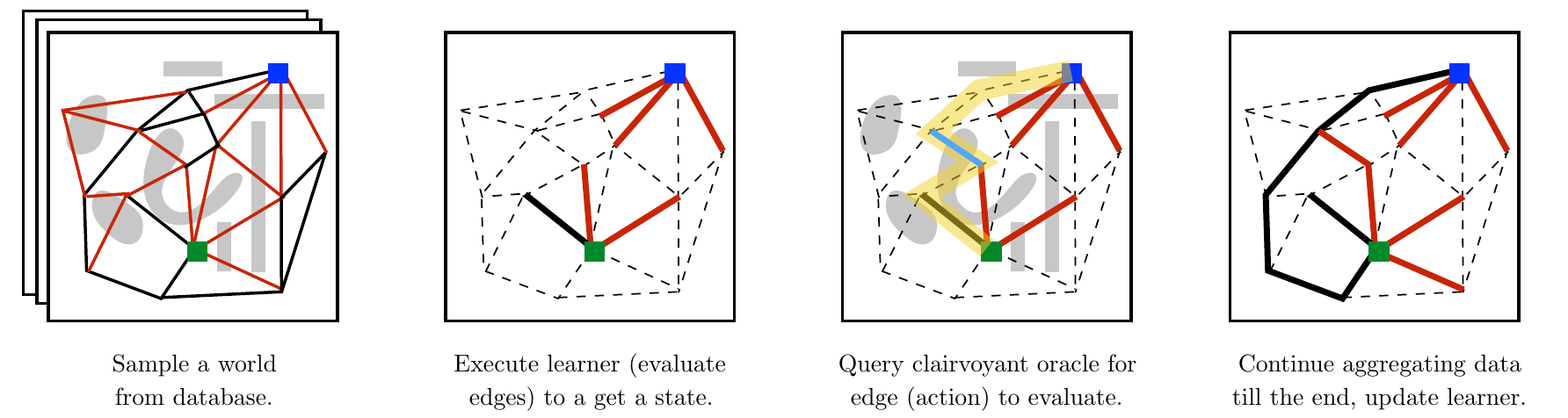}
    \caption{ Overview of \algName - a training procedure for a \selector to select edges to evaluate in the \lazysp framework. 
    In each training iteration, a world map $\world$ is sampled. The learner is executed upto a time step to get a state $s_t$ which is the set of edges evaluated and their outcomes. The learner has to decide which edge to evaluate on the current shortest path. It extracts features from every edge - we use a set of baseline heuristic values as features. The \selector asks a clairvoyant oracle selector (which has full knowledge of the world) which edge to evaluate. This is then added to a classification dataset and the learner is updated. This process is repeated over several iterations. \fullFigGap}
  \label{fig:illustration_algorithm}
\end{figure*}

Interestingly, if we were to reveal the status of all the edges during training, we can conceive of a \emph{clairvoyant oracle}~\citep{choudhury2017data} that can select the optimal sequence of edges to invalidate. In fact, we show that the oracular selector is equivalent to \emph{set cover}, for which greedy approximations exist. By imitating clairvoyant oracles~\citep{choudhury2017data}, we can drastically cut down on exploration and focus learning on a small, relevant portion of the state space~\cite{sun2017deeply}. This leads to a key insight: use imitation learning to quickly bootstrap the selector to match oracular performance.  % our key insight - 
%\begin{quote}
%Bootstrap selector learning by imitating oracular selectors.
%\end{quote}
%
We propose a new algorithm, \algName, that deploys an interactive imitation learning framework~\citep{ross2011reduction} to train the edge selector (\figref{fig:illustration_algorithm}). At every iteration, it samples a world (validity status for all edges) and executes the learner. At every timestep, it queries the clairvoyant oracle associated with the world to select an edge to evaluate. This can be viewed as a classification problem where the goal is to map features extracted from edges to the edge selected by the oracle. This datapoint is aggregated with past data, which is then used to update the learner. 

In summary, our main contributions are:
\begin{enumerate}
	\item We map edge selection in lazy search to an MDP (\sref{sec:problem_formulation}) and solve it for small graphs (\sref{sec:challenge}).
	\item We show that larger MDPs, can be efficiently solved by imitating clairvoyant oracles (\sref{sec:approach}).
	\item We show that the learned policy  can outperform competitive baselines on a wide range of datasets (\sref{sec:experiments}).
\end{enumerate}

\section{Problem Formulation}
\label{sec:problem_formulation}

The overall objective is to design an algorithm that can solve the Shortest Path (SP) problem while minimizing the number of edges evaluated.

\subsection{The Shortest Path (SP) Problem}
\label{sec:problem_formulation:sp}

Let $\explicitGraph = \pair{\vertexSet}{\edgeSet}$ be an explicit graph where $\vertexSet$ denotes the set of vertices and $\edgeSet$ the set of edges. Given a start and goal vertex $\pair{\start}{\goal} \in \vertexSet$, a path $\path$ is represented as a sequence of vertices $\seq{\vertex}{l}$ such that $\vertex_1 = \start, \vertex_l = \goal, \forall i,~\pair{\vertex_i}{\vertex_{i+1}} \in \edgeSet$. 
We define a \emph{world} $\world: \edgeSet \rightarrow \{0,1\}$ as a mapping from edges to valid ($1$) or invalid ($0$). A path is said to be \emph{feasible} if all edges are valid, i.e. $\forall \edge \in \path, \world(\edge) = 1$. 
Let $\length: \edgeSet \rightarrow \real^+$ be the length of an edge. The length of a path is the sum of edge lengths, i.e. $\length(\path) = \sum_{\edge \in \path} \length(\edge)$. The objective of the SP problem is the find the shortest feasible path:
\begin{equation}
    \min_{\path} \; \length(\path) \suchthat{ \forall \edge\in\path, \world(\edge) = 1 }
\end{equation}
We now define a family of shortest path algorithms. Given a SP problem, the algorithms evaluate a set of edges $\evalEdges \subset \edgeSet$ (verify if they are valid) and return a path $\path^*$ upon halting. Two conditions must be met:
\begin{enumerate}
    \item The returned path $\path^*$ is verified to be feasible, i.e. $\forall \edge \in \path^*,~\edge \in \evalEdges,~\world(\edge) = 1$
    \item All paths shorter than $\path^*$ are verified to be infeasible, i.e. $\forall \path_i,~\length(\path_i) \leq \length(\path^*),~\exists \edge \in \path_i,~\edge \in \evalEdges,~\world(\edge) = 0 $
\end{enumerate}

\subsection{The Lazy Shortest Path (\lazysp) Framework}
\label{sec:problem_formulation:lazysp}
We are interested in shortest path algorithms that minimize the number of evaluated edges $\abs{\evalEdges}$.%
\footnote{The framework can be extended to handle non-uniform evaluation cost as well}
These are \emph{lazy} algorithms, i.e. they seek to defer the evaluation of an edge as much as possible. 
When this laziness is taken to the limit, one arrives at the \emph{Lazy Shortest Path} (\lazysp) class of algorithms. 
Under a set of assumptions, this framework can be shown to contain the optimally lazy algorithm~\citep{haghtalab2017provable}. 

\algoref{alg:lsp} describes the \lazysp framework. The algorithm maintains a set of evaluated edges that are valid $\edgesValid$ and invalid $\edgesInvalid$. At every iteration, the algorithm lazily finds the shortest path $\path$ on the potentially valid graph $\explicitGraph = \pair{\vertexSet}{\edgeSet \setminus \edgesInvalid}$ \emph{without evaluating any new edges} (\algolineref{alg:lsp:sp}). It then calls a function, \selector, to select an edge $\edge$ from this path $\path$ (\algolineref{alg:lsp:select}). Depending on the outcome, this edge is added to either $\edgesValid$ or $\edgesInvalid$. This process continues until the conditions in \sref{sec:problem_formulation:sp} are satisfied, i.e. the shortest feasible path is found. 

\begin{algorithm}[tb]
%\setstretch{1.2}
\SetAlgoLined
\caption{\lazysp \label{alg:lsp}}
\SetKwInOut{Input}{Input}
% \vspace{1mm}
\SetKwInOut{Parameter}{Parameter}
\SetKwInOut{Output}{Output}
\Input{$\text{Graph } \explicitGraph, \text{ start } \start{}, \text{ goal } \goal{}, \text{ world } \world$}
\Parameter{$\selector$}
\Output{$\text{Path}\ \path^*, \text{ evaluated edges } \evalEdges{}$}
\vspace{2mm}
$\edgesValid \gets \emptyset$ \Comment{Valid evaluated edges} \\
$\edgesInvalid \gets \emptyset$ \Comment{Invaid evaluated edges} \\
\vspace{1mm}
\Repeat{feasible path found $\text{s.t.} \; \forall \edge \in \path, \edge \in \edgesValid$}
{
$\path \gets \textsc{ShortestPath}(\edgeSet \setminus \edgesInvalid)$ \label{alg:lsp:sp}\\
$\edge \gets \selector(\path, \edgesValid, \edgesInvalid)$ \Comment{Select edge on $\path$} \label{alg:lsp:select}\\
\uIf{$\world(\edge)\neq 0$}
{
    $\edgesValid \gets \edgesValid \cup \{ \edge \}$
}
\Else
{
    $\edgesInvalid \gets \edgesInvalid \cup \{ \edge \}$
}
% $\LazyTree \gets $ \textsc{ExtendTree~(\Event, \LazyTree)} \Comment{Add \Vexp{}} \\
% \vspace{0.5mm}
% $\Subpath \gets $ \textsc{GetShortestPathToLeaf~(\LazyTree)} \\
% \vspace{0.5mm}
% \textsc{EvaluateEdge~(\Selector, \Subpath)}  \Comment{Add \Eeval{}} \\
}
\KwRet\ \{$\path^* \gets \path$, $\evalEdges \gets \edgesValid \cup \edgesInvalid$\};
\end{algorithm}

The algorithm has one free parameter - the \selector function. The only requirement for a valid \selector is to select an edge on the path. As shown in \citep{dellin2016unifying}, one can design a range of selectors such as:
\begin{enumerate}
    \item \selectorForward: select the first unevaluated edge $\edge \in \path$. Effective if invalid edges are near the start.
    \item \selectorBackward: select the last unevaluated edge $\edge \in \path$. Effective if invalid edges are near the goal.
    \item \selectorAlternate: alternates between first and last edge. This approach hedges its bets between start and goal.
    \item \selectorFailFast: selects the least likely edge $\edge \in \path$ to be valid based on prior data. 
    \item \selectorPostFailFast: selects the least likely edge $\edge \in \path$ to be valid using a Bayesian posterior based on edges checked so far.
\end{enumerate}

While these baseline selectors are very effective in practice, their performance, i.e. the number of edges evaluated $\abs{\evalEdges}$ depends on the underlying world $\world$ which dictates which edges are invalid. Hence the goal is to compute a good \selector that is effective given a \emph{distribution of worlds}, $P(\world)$. We formalize this as follows
%should select edges that are not only likely to be invalid but also have a lot of paths flowing through them. 

\begin{problem}[Optimal Selector Problem] \label{prob:opt_select}
Let the edges evaluated by \selector on world $\world$ be denoted by $\evalEdges(\world, \selector)$. Given a distribution of worlds, $P(\world)$, find a \selector that minimizes the expected number of evaluated edges, i.e. $\min \expect{\world \sim P(\world)}{\abs{\evalEdges(\world, \selector)}}$
\end{problem}

Problem~\ref{prob:opt_select} is a sequential decision making problem, i.e. decisions made by the selector in one iteration (edge selected) affects the input to the selector in the next iteration (shortest path). We show how to formally handle this in the next section. It's interesting to note that Problem~\ref{prob:opt_select} can be solved optimally under certain strong assumptions (See supplementary for details\footnote{Supplementary material can be found at: \href{http://bit.ly/2Em6Meu}{http://bit.ly/2Em6Meu}}).

\subsection{Mapping the Optimal Selector Problem to an MDP}
\label{sec:problem_formulation:mdp}
We map Problem~\ref{prob:opt_select} to a Markov Decision Process (MDP) $\langle \stateSpace, \actionSpace, \transFn, \rewardFn \rangle$ as follows:

\subsubsection*{State Space}
The state $\state = (\edgesValid, \edgesInvalid)$ is the set of evaluated valid edges $\edgesValid$ and evaluated invalid edges $\edgesInvalid$. This can be represented by a vector of size $\abs{\edgeSet}$, each element being one of $\{-1, 0, 1\}$ - unevaluated, evaluated invalid, and evaluated valid respectively.
For simplicity, we assume that the explicit graph $\explicitGraph = (\vertexSet, \edgeSet)$ is fixed.\footnote{We can handle a varying graph by adding it to the state space.} 

Since each $\edge \in \edgeSet$ can be in one of $3$ sets, the cardinality of the state space is $\card{S} = 3^{\card{\edgeSet}}$.

The MDP has an absorbing goal state set $\stateAbs \subset \stateSpace$ which is a set of states where all the edges on the current shortest path are evaluated to be valid, i.e. 
\begin{equation}
    \stateAbs = \setst{ ( \edgesValid, \edgesInvalid ) }{ \forall \edge \in \textsc{ShortestPath}(\edgeSet \setminus \edgesInvalid), \edge \in \edgesValid}
\end{equation}
 
\subsubsection*{Action Set}
The action set $\actionSpace(\state)$ is the set of unevaluated edges on the current shortest path, i.e.
\begin{equation}
    \actionSpace(\state) = \{ \edge \in \textsc{ShortestPath}(\edgeSet \setminus \edgesInvalid), \edge \notin \{ \edgesValid \cup \edgesInvalid \} \}
\end{equation}

\subsubsection*{Transition Function}
Given a world $\world$, the transition function is deterministic $\state' = \Gamma(\state, \action, \world)$:
\begin{equation}
    \Gamma(\state, \action, \world) =   \begin{cases} 
   (\edgesValid \cup \{\edge\}, \edgesInvalid)           & \text{if } \world(\edge) = 1 \\
   (\edgesValid, \edgesInvalid \cup \{ \edge \})         & \text{if } \world(\edge) = 0
  \end{cases}
\end{equation}

Since $\world$ is latent and distributed according to $P(\world)$, we have a stochastic transition function $\transFn(\state, \action, \state') = \sum_{\world} P(\world) \Ind(\state = \Gamma(\state, \action, \world))$.

\subsubsection*{Reward Function}
The reward function penalizes every state other than the absorbing goal state $\stateAbs$, i.e.
\begin{equation}
    \rewardFn(\state, \action) =   \begin{cases} 
   0             & \text{if } \state \in \stateAbs  \\
   -1            & \text{otherwise } 
  \end{cases}
\end{equation}

% !TEX root = ../main.tex

\section{Challenges in Solving the MDP}
\label{sec:challenge}

In this section, we examine tiny graphs and show that even for such problems, a choice of world distributions where edges are correlated can affect \selector choices. However, by solving the MDP using tabular Q-learning we can automatically recover the optimal \selector. 

\subsection{Experimental setup}
We train selectors on two different graphs and corresponding distribution of worlds $P(\world)$. 

\begin{figure}[!t]
\centering
\includegraphics[width=\columnwidth]{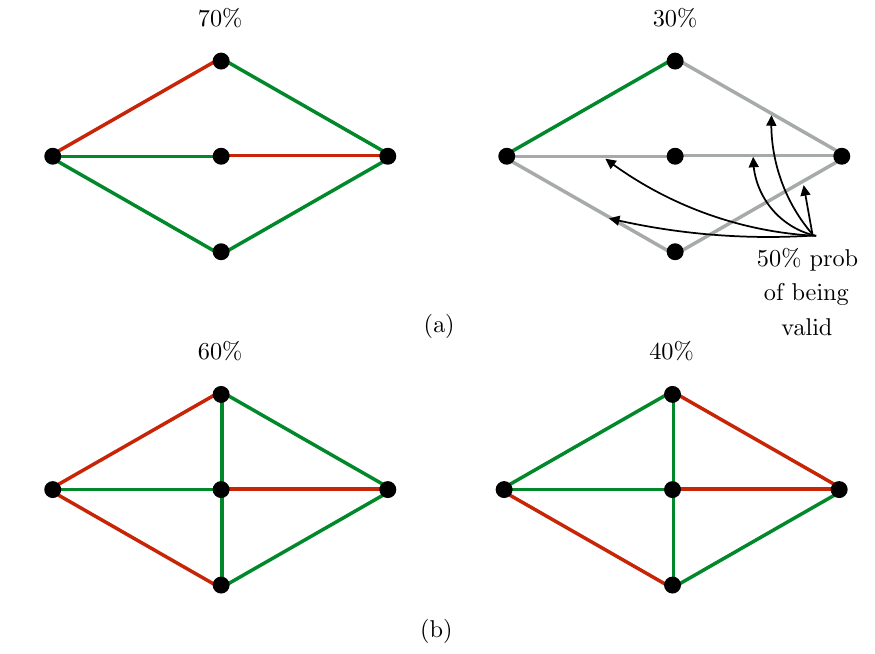}
\caption{Distribution over worlds for (a) Environment 1 and (b) Environment 2. The goal is to find a path from left to right. Edges are valid (green) or invalid (red) \fullFigGap}
\label{fig:simple_env}
\end{figure}

\subsubsection*{Environment 1} 
Fig.~\ref{fig:simple_env}(a) illustrates the distribution of Environment 1. The graph has $6$ edges. With $70 \%$ probability, $\mathtt{top\_left}$ edge is invalid. If $\mathtt{top\_left}$ is invalid, then $\mathtt{middle\_right}$ is always invalid. If $\mathtt{top\_left}$ is valid, then with $50\%$ probability, $\mathtt{top\_right}$ is invalid plus any one of remaining four are invalid. 

The optimal policy is to check $\mathtt{top\_left}$ edge first. 
\begin{itemize}
    \item[--] \emph{If invalid}, check $\mathtt{middle\_right}$ (which is necessarily invalid) and check bottom two edges which are feasible. This amounts to $4$ evaluated edges.
    \item[--] \emph{If valid}, check other edges in order as they all have $50\%$ probability of being valid.
\end{itemize}

\subsubsection*{Environment 2} 
Fig.~\ref{fig:simple_env}(b) illustrates the distribution of Environment 2. The graph has $8$ edges. 
With $60 \%$ of the time $\mathtt{top\_left}$, $\mathtt{middle\_right}$ and $\mathtt{bottom\_left}$ are invalid. Else, $\mathtt{top\_right}$ and $\mathtt{middle\_right}$ are invalid. 
Intuitively, $60 \%$ of the time, \textsc{SelectAlternate} is optimal and $40 \%$ of the time, \textsc{SelectBackward} is the best. 

\subsection{Solving the MDP via Q-learning}

We apply tabular Q-learning~\cite{watkins1992q} to compute the optimal value $Q^*(\state, \action)$. Broadly speaking, the algorithm uses an $\epsilon-$greedy algorithm to visit states, gather rewards, and perform Bellman backups to update the value function. Environment 1 has $729$ states, Environment 2 has $6561$ states. The learning parameters are shown in Table~\ref{tab:q_learning_params}. 

\begin{figure}[!t]
    \centering
    \captionsetup[subfigure]{justification=centering}
    \begin{subfigure}[t]{0.8\columnwidth}
        \includegraphics[height=1.0in]{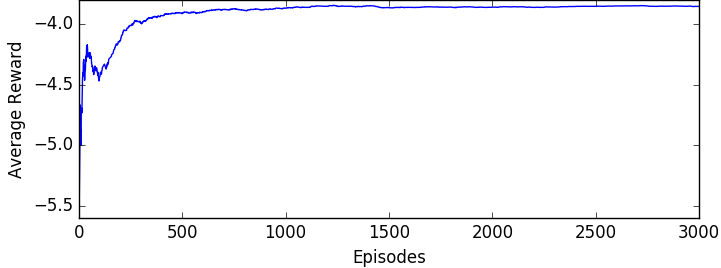}
        \caption{Environment 1 (3000 train episodes)}
    \end{subfigure}
   \hspace{15mm}
    \begin{subfigure}[t]{0.8\columnwidth}
        \includegraphics[height=1.0in]{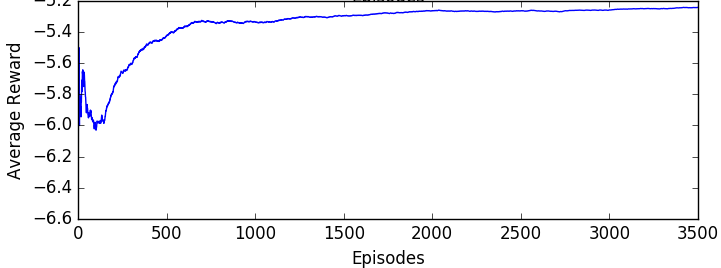}
        \caption{Environment 2 (3500 train episodes)}
    \end{subfigure}
    \caption{ Average reward per epsiode of Tabular Q-learning. \fullFigGap}
    \label{fig:tab_q_results}
\end{figure}

\figref{fig:tab_q_results} shows the average reward during training for Q-learning. Environment 1 converges after $\approx 1000$ episodes, environment 2 after $\approx 3000$ episodes. Table~\ref{tab:experimental_results} shows a comparison of Q-learning with other heuristic baselines in terms of average reward on a validation dataset of $1000$ problems. In Environment 1, the learner discovers the optimal policy. Interestingly, \selectorAlternate also achieves this result since the correlated edges are alternating.  In Environment 2, the learner has a clear margin as compared to heuristic baselines, all of which are vulnerable to one of the modes.

This shows that, even on such small graphs, it is possible to create an environment where heuristic baselines fail. The fact that the learner can recover optimal policies is promising. 

\begin{table}[!t]
\centering
\caption{Q-learning parameters.}
\begin{tabulary}{\columnwidth}{LCC}\toprule
 {\bf Parameter} & {\bf Environment 1} & {\bf Environment 2} \\ \midrule
 Number of episodes & $3000$ & $3500$ \\
 Exploration episodes & $100$ & $150$ \\
 $\epsilon_0$  & $1$ & $1$ \\
 Discount factor & $1$ & $1$ \\
 Learning rate & $0.5$ & $0.5$ \\
 \bottomrule
\end{tabulary}
\label{tab:q_learning_params}
\end{table}

\begin{table}[t!]
\centering
\caption{Average reward after 1000 test episodes.}
\begin{tabulary}{\columnwidth}{LCC}\toprule
 {\bf Method}  & {\bf Environment 1} & {\bf Environment 2} \\ \midrule
 Tabular Q-learning  & $\cmark -3.85$ & $\cmark-5.24$\\
 \selectorForward & $-4.54$ & $-6.00$ \\
 \selectorBackward  & $-4.42$ & $-5.79$ \\
 \selectorAlternate & $-3.86$ & $-6.00$\\
 \selectorRandom & $-4.48$ & $-5.90$ \\
 \bottomrule
\end{tabulary}
\label{tab:experimental_results}
\end{table}

\subsection{Challenges on scaling to larger graphs}

While we can solve the MDP for tiny graphs, we run into a number of problems as we try to scale to larger graphs:

\paragraph{Exponentially large state space}
The size of the state space is $\card{S} = 3^{\card{\edgeSet}}$. This leads to exponentially slower convergence rates as the size of the graph increases. Even if we could manage to visit only the relevant portion of this space, this approach would not generalize across graphs. 

\paragraph{Convergence issues with approximate value iteration}
We can scale to large graphs if we use a function approximator. In this case, we have to featurize $(\state, \action)$ as a vector $f$, i.e. we are trying to approximate $Q(s,a) \approx Q(f)$. Fortunately, we have a set of baseline heuristics~\ref{sec:problem_formulation:lazysp} that can be used as a feature vector. This choice allows us to potentially improve upon baselines and easily switch between problem domains.

We run into another problem - approximate value iteration is not guaranteed to converge~\citep{gordon1995stable}. This is exaggerated in our case where $f$ is a set of baseline heuristics that may not retain the same information content as the state $\state$. Hence multiple states map to the same feature $f$, which leads to oscillations and local minima.

\paragraph{Sparse rewards}
Every state gets a penalization except the absorbing state, i.e. rewards are sparse. Because we are using a function approximator, updates to $Q(f)$ for reaching the goal state are overridden by updates due to $-1$ penalization.

% !TEX root = ../main.tex

\section{Approach}
\label{sec:approach}

Our approach, \algName (\algFullName), is to imitate clairvoyant oracles that can show how to evaluate edges optimally given full knowledge of the MDP at training time. To deal with distribution mismatch between oracle and learner, we use established techniques for iterative supervised learning.

\subsection{Optimistic Value Estimate using a Clairvoyant Oracle}
\label{sec:approach:oracle}
Consider the situation where the world $\world$ is fully known to the selector, i.e. the $0/1$ status of all edges are known. The selector can then judiciously select edges that are not only invalid, but eliminate paths quickly. We call such a selector a \emph{clairvoyant oracle}. We show that the optimal clairvoyant oracle, that evaluates the minimal number of edges, is the solution to a set cover problem.
\begin{theorem}[Clairvoyant Oracle as Set Cover]
\label{thm:set_cover}
Let $\state = (\edgesValid, \edgesInvalid)$ be a state. Let $V^*(\state, \world)$ be the optimal state action value when the world $\world$ is known. Then $V^*(\state, \world)$ is the solution to the following set cover problem
\begin{equation}
\begin{aligned}
\label{eq:set_cover}
-\minprob{\substack{\evalEdges \subset \setst{\edge \in \edgeSet}{\world(\edge)=0}} } & \abs{\evalEdges} \\
\mathrm{s.t.} \quad & \forall \path,\; \length(\path) \leq \length(\path^*), \path \cap \edgesInvalid = \emptyset, \\
		   & \hphantom{\forall \path,\;} \path \cap \evalEdges \neq \emptyset
\end{aligned}
\end{equation}
\end{theorem}
where $\path^*$ is the shortest feasible path for world $\world$.
\begin{proof}
(Sketch) Let $\pathSet = \{\path_1, \dots, \path_n\}$ be the set of paths that satisfy the constraints of~\eref{eq:set_cover}
\begin{enumerate}
	\item Shorter than $\path^*$, i.e. $\length(\path_i) \leq \length(\path^*)$
	\item Paths are not yet invalidated i.e. $\path \cap \edgesInvalid = \emptyset$
\end{enumerate}

Let $\setst{\edge \in \edgeSet}{\world(\edge)=0}$ be the set of invalid edges. Each edge $\edge$ covers a path $\path_i \in \pathSet$ if $\edge \in \path_i$. 
We define a cover as a set of edges $\evalEdges$ that covers all paths in $\pathSet$, i.e. $\path_i \cap \evalEdges \neq \emptyset$.

If we select a min cover, i.e. $\min \abs{\evalEdges}$ then all shorter paths will be eliminated. Hence this is equal to the optimal value $-V^*(\state, \world)$.
\end{proof}
%\boots{I think you should explain the math above with words.}

Theorem~\ref{thm:set_cover} says that given a world and a state of the search, the clairvoyant oracle selects the minimum set of invalid edges to eliminate paths shorter than the shortest feasible path. 

Let $\policyOracle(\state,\world)$ be the corresponding oracle policy. We note that the optimal clairvoyant oracle can be used to derive an upper bound for the optimal value
\begin{equation}
\label{eq:upper_bound}
	Q^*(\state, \action) \leq  Q^{\policyOracle}(s,a) = \sum_{\world} P(\world | \state) Q^{\policyOracle}(s,a,\world)
\end{equation}
where $P(\world | \state)$ is the posterior distribution over worlds given state and $Q^{\policyOracle}(s,a,\world)$ is the value of executing action $a$ in state $s$ and subsequently rolling-out the oracle. Hence this upper bound can be used for learning. 

\subsection{Approximating the Clairvoyant Oracle}
\label{sec:approach:approx_oracle}

\begin{algorithm}[!t]
	\caption{\textsc{Approximate Clairvoyant Oracle}   \label{alg:approximate_oracle}}
	\SetKwInOut{Input}{Input}
% \vspace{1mm}
\SetKwInOut{Output}{Output}
\Input{$\text{State } \state = (\edgesValid, \edgesInvalid), \text{ world } \world$}
\Output{$\text{Action } \action$}
	Compute shortest path $\hat{\path} = \textsc{ShortestPath}(\edgeSet \setminus \edgesInvalid)$ \\
	$\Delta \gets 0_{\abs{\edgeSet} \times 1}$\\
	\For{$\edge \in \hat{\path}, \world(\edge) = 0 $}
	{
		$\Delta(\edge) \gets \length(\textsc{ShortestPath}(\edgeSet \setminus \{ \edgesInvalid \cup \{ \edge \} \} )) - \length(\hat{\path})$\\
	}
	\KwRet\ Action $a = \argmaxprob{\edge \in \hat{\path}} \Delta(\edge)$;
\end{algorithm}

Since set cover is NP-Hard, we have to approximately solve \eref{eq:set_cover}. Fortunately, a greedy approximation exists which is near-optimal. The greedy algorithm iterates over the following rule:
%selects the edge which covers the maximum number of shorter paths to eliminate according to the following selection rule:
\begin{equation}
\begin{aligned}
\label{eq:greedy_set_cover}
 	&\edge_i = \argmaxprob{\edge \in \edgeSet, \world(\edge)=0} \;\abs{ \setst{\path}{\length(\path) \leq \length(\path^*), \; \path \cap \edgesInvalid = \emptyset, \; \edge \in \path} } \\
 	&\evalEdges \gets \evalEdges \cup \{ \edge_i \}
\end{aligned}
\end{equation} 
%\boots{Again, I think you should walk the reader through the above selection rule be explaining what is happening in the text.}
The approach greedily selects an invalid edge that covers the maximum number of shorter paths, which have not yet been eliminated. This greedy process is repeated until all paths are eliminated.

There are two practical problems with computing such an oracle. First, enumerating all shorter paths $\setst{\path}{\length(\path) \leq \length(\path^*)}$ is expensive, even at train time. Second, if we simply wish to query the oracle for which edge to select on the current shortest path $\hat{\path} = \textsc{ShortestPath}(\edgeSet \setminus \edgesInvalid)$, it has to execute \eref{eq:greedy_set_cover} potentially multiple times before such an edge is discovered - which also can be expensive. Hence we perform a double approximation.

The first approximation to \eref{eq:greedy_set_cover} is to constrain the oracle to only select an edge on the current shortest path $\hat{\path} = \textsc{ShortestPath}(\edgeSet \setminus \edgesInvalid)$ \vspace{-2mm}
\begin{equation}
\label{eq:constr_set_cover}
\approx \argmaxprob{\edge \in \hat{\path}, \; \world(\edge)=0} \;\abs{ \setst{\path}{\length(\path) \leq \length(\path^*), \; \path \cap \edgesInvalid = \emptyset, \; \edge \in \path} } 
\end{equation} 

The second approximation to \eref{eq:constr_set_cover} is to replace the number of paths covered with the marginal gain in path length on invalidating an edge. 
\begin{equation}
\label{eq:length_surrogate}
\approx \argmaxprob{\edge \in \hat{\path}, \; \world(\edge)=0} \; \length(\textsc{ShortestPath}(\edgeSet \setminus \{ \edgesInvalid \cup \{\edge\} \} )) - \length(\hat{\path})
\end{equation} 

Alg.~\ref{alg:approximate_oracle} summarizes this approximate clairvoyant oracle. 

\subsection{Bootstrapping with Imitation Learning}
\label{sec:approach:imitation}

% While the clairvoyant oracle gives us a good upper bound, we cannot actually use this oracle at test time to compute the value estimate (\ref{eq:upper_bound}) to select an action. This requires us to enumerate all plausible worlds and run the oracle which is prohibitively expensive. Instead, we want a policy $\policy(\state)$ that directly maps state to action.

Imitation learning is a principled way to use the clairvoyant oracle $\policyOracle(\state, \world)$ to assist in training the learner $\policy(\state)$. In our case, we can use the oracle action value $Q^{\policyOracle}(\state, \action)$ as a target for our learner as follows:
\begin{equation}
\label{eq:aggrevate}
\argmaxprob{\policy \in \policySpace}\;  \expect{\state \sim d_{\policy}(\state)}{Q^{\policyOracle}(\state, \policy(\state))} \\
\end{equation}
where $d_{\policy}(\state)$ is the distribution of states. Note that this is now a classification problem since the labels are provided by the oracle. However the distribution $d_{\policy}$ depends on the learner's $\policy$. \citet{ross2014reinforcement} show that this type of imitation learning problem can be reduced to interactive supervised learning.

We simplify further. Computing the oracle value requires rolling out the oracle until termination. We empirically found this to significantly slow down training time. Instead, we train the policy to directly predict the action that is selected by the oracle. This is the same as (\ref{eq:aggrevate}) but with a $0/1$ loss~\citep{ross2011reduction} -
\begin{equation}
\label{eq:dagger}
\argmaxprob{\policy \in \policySpace}\;  \expect{\state \sim d_{\policy}(\state)}{ \Ind(\policy(\state) = \policyOracle(\state, \world)) } \\
\end{equation}

We justify this simplification by first showing that maximizing action value is same as maximizing the advantage $Q^{\policyOracle}(\state, \action) - V^{\policyOracle}(\state)$. Since all the rewards are $-1$, the advantage can be lower bounded by the $0/1$ loss. We summarize this as follows:
\begin{equation}
\begin{aligned}
&\maxprob{\policy \in \policySpace}\;  \expect{\state \sim d_{\policy}(\state)}{Q^{\policyOracle}(\state, \policy(\state))} \\
= \quad & \maxprob{\policy \in \policySpace} \; \expect{\state \sim d_{\policy}(\state)}{Q^{\policyOracle}(\state, \policy(\state)) - V^{\policyOracle}(\state)} \\
\geq \quad & \maxprob{\policy \in \policySpace} \; \expect{\state \sim d_{\policy}(\state)}{ \Ind(\policy(\state) = \policyOracle(\state, \world)) - 1 } \\
\end{aligned}
\end{equation}

Finally, we do not use the exact clairvoyant oracle but rather an approximation (Section~\ref{sec:approach:approx_oracle}). In other words, there can exist policies $\policy \in \policySpace$ that outperform the oracle. In such a case, one can potentially apply policy improvement after imitation learning. However, we leave the exploration of this direction to future work.

\subsection{Algorithm}
\label{sec:approach:algorithm}

\begin{algorithm}[!t]
	\caption{\algName   \label{alg:lsp_learn}}
	\SetKwInOut{Input}{Input}
% \vspace{1mm}
\SetKwInOut{Parameter}{Parameter}
\SetKwInOut{Output}{Output}
\Input{$\text{World distribution } P(\world), \text{ oracle } \policyOracle$}
\Parameter{$\text{Iter } N, \text{roll-in policy } \policyRoll, \text{ mixing } \{\mixParam_i\}_{i=1}^N$}
\Output{$\text{Policy } \policyLearn$}
	Initialize $\dataset \leftarrow \emptyset,\; \policyLearn_{1}$ to any policy in $\policySpace$ \\ \label{lst:line:}
	\For{$i = 1, \ldots, N$}
	{   Initialize sub-dataset $\dataset_{i} \leftarrow \emptyset$ \\
 		Let mixture policy be $\policyMix =\;\mixParam_{i}\policyRoll\; + \; (1 \; - \;\mixParam_{i})\policyLearn_{i} $ \\
 		\For{$j = 1,\ldots,m$}
 		{   
 			Sample $\world \sim P(\world)$; \\  
 			Rollin $\policyMix$ to get state trajectory $\{ \state_t \}_{t=1}^T$ \\
 			Invoke oracle to get $\action_t = \policyOracle(s_t, \world)$ \\
		    $\dataset_i \gets \dataset_{i} \cup \{ \left( \state_t, \action_t \right) \}_{t=1}^T$ \; 
 	    }
 	    Aggregate data $\dataset \leftarrow \dataset \cup \dataset_{i}$; \\
 	    Train classifier $\policyLearn_{i+1}$  on $\dataset$;\\  
	}
	\KwRet\ Best $\policyLearn$ on validation;
\end{algorithm}

The problem in (\ref{eq:dagger}) is a non-\emph{i.i.d} classification problem - the goal is to select the same action the oracle would select on the \emph{on policy distribution of learner}. \citet{ross2011reduction} proposed an algorithm, \daggerAlg, to exactly solve such problems. 

Alg.~\ref{alg:lsp_learn}, describes the $\algName$ framework which iteratively trains a sequence of policies $\seq{\policyLearn}{N}$. At every iteration $i$, we collect a dataset $\dataset_i$  by executing $m$ different episodes. In every episode, we sample a world $\world$ which already has every edge evaluated. We then roll-in a policy (execute a selector) which is a mixture $\policyMix$ that blends the learner's current policy, $\policyLearn_{i}$ and a base roll-in policy $\policyRoll$ using blending parameter $\mixParam_{i}$. At every time step $t$, we query the clairvoyant oracle with state $\state_t$ to receive an action $\action_t$. We use the approximate oracle in Alg.~\ref{alg:approximate_oracle}. We then extract a feature vector $f$ from all $(\state_t, \action)$ tuples and create a classification datapoint. We add this datapoint to the dataset $\dataset_i$. At the end of $m$ episodes, this data is then \emph{aggregated with the existing dataset} $\dataset$. A new classifier $\policyLearn_{i+1}$ is trained on the aggregated data. At the end of $N$ iterations, the algorithm returns the best performing policy on a set of held-out validation environments. 

We have two algorithms based on the choice of $\policyRoll$:
\begin{enumerate}
	\item \algName: We set $\policyRoll = \policyOracle$. This is the default mode of \daggerAlg. This uses the oracle state distribution to stabilize learning initially. 
	\item \algHeuristic: We set $\policyRoll$ to be the best performing heuristic on training as defined in \sref{sec:problem_formulation:lazysp}. This uses a heuristic state distribution to stabilize learning. Since the heuristic is realizable, it can have a stabilizing effect on datasets where the oracle is far from realizable.
\end{enumerate}

We inherit the performance guarantees of \daggerAlg~\citep{ross2011reduction}, which bounds the performance gap with respect to the best policy in the policy class. 

% !TEX root = ../main.tex

\section{Experiments}
\label{sec:experiments}

\subsection{Experimental Setup}
%We use datasets from \cite{choudhury2017active} for our experiments. We have developed OpenAI Gym \citep{brockman2016gym} environments corresponding to the shortest path MDP where the true edge statuses are dictated by worlds sampled from these datasets. \footnote{Code for our implementation will be open-sourced.}
We use datasets from \cite{choudhury2017active} in our experiments. The 2D datasets contain graphs with approximately 1600-5000 edges and varied obstacle distributions. The two 7D datasets involve a robot arm planning for a reaching task in clutter with large graphs containing 33286 edges.

\subsubsection*{Learning Details}
We only consider policies that are a linear combination of a minimal set of features, where each feature is a different motion planning heuristic. The features we consider are: 
\begin{enumerate}
    \item \featPrior - the prior probability of an edge being invalid calculated over the training dataset.
    \item \featPost - the posterior probability of an edge being invalid given collision checks done thus far (See supplementary for details). %\ref{appendix:posterior} for details). \footnote{Supplementary material can be found at: \href{http://bit.ly/2Em6Meu}{http://bit.ly/2Em6Meu}}
    \item \featIndex - score ranging from 1 (first unchecked edge) to 0 (last unchecked edge).
    \item \featDeltaLength - hallucinate that an edge is invalid, then calculate the difference in length of new shortest path compared with the current shortest path.
    \item \featDeltaEval - hallucinate that an edge is invalid, the calculate the fraction of unevaluated edges on the new shortest path.
    \item \featPDL - calculated as \featPost $\times$ \featDeltaLength, it weighs the \featDeltaLength of an edge with the probability of it being invalid and is effective in practice (Table \ref{tab:benchmark_results}).
\end{enumerate}  

\subsection{Baselines}
We compare our approach to common heuristics used in \lazysp as described in Section \ref{sec:problem_formulation:lazysp}. We also analyze the improvement in performance as compared to vanilla behavior cloning of the oracle and reinforcement learning from scratch. 

\subsection{Analysis of Overall Performance}
\begin{observation}
\algName has consistently strong performance across different datasets. 
\end{observation}
Table \ref{tab:benchmark_results} shows that \algName is able to learn policies competitive with other motion planning heuristics. No other heuristic has as consistent a performance across datasets.

\begin{figure*}[!t]
\centering
    \includegraphics[width=\textwidth]{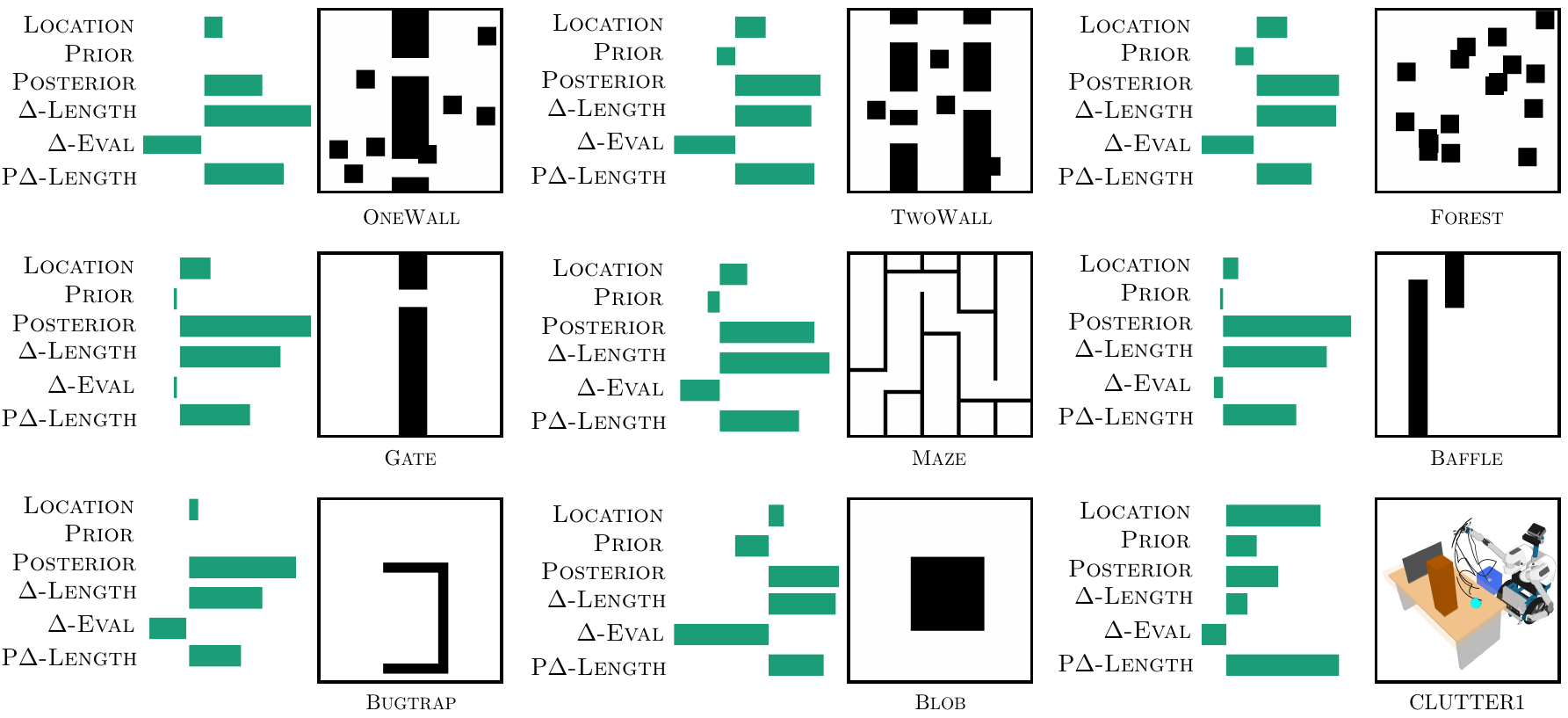}
\caption{ Weight bins depicting relative importance of each feature learned by the learner. \algName focuses on edges that are highly likely to be invalid and have high measure of centrality. \fullFigGap}
\label{fig:weights_learner} 
\end{figure*}

\begin{observation}
The learner focuses collision checking on edges that are highly likely to be invalid and have a high measure of centrality.  
\end{observation} 
\figref{fig:weights_learner} shows the activation of different features across datasets. The learner places high importance on \featPost, \featDeltaLength and \featPDL. \featPost is an approximate likelihood of an edge being invalid and \featDeltaLength is an approximate measure of centrality i.e. edges with large \featDeltaLength have large number of paths passing through them (Note that the converse may not always apply).

%On datasets with concentrated obstacles eg. \textsc{blob}, the learner up-weights \featPrior\; whereas, when obstacles are spread out eg. \textsc{forest}, the learner up-weights \featIndex \; and \featDeltaLength, to search for bottleneck edges.

\begin{observation}
On datasets with strong correlations among edges, heuristics that take obstacle distribution into account outperform uninformed heuristics, and \algName is able to learn significantly better policies than uninformed heuristics.
\end{observation}

\begin{figure}[!t]
\centering
    \includegraphics[width=\columnwidth]{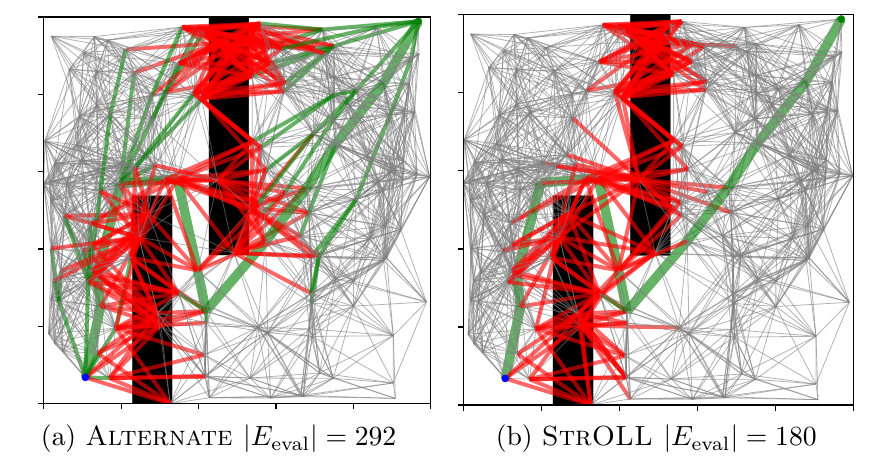}
\caption{Edges evaluated (green valid, red invalid) on a world from \textsc{Baffle}. (a) \selectorAlternate evaluates several valid edges (b) \algName evaluates many fewer edges, all of which are invalid and eliminate a large number of paths. \fullFigGap}
 \label{fig:edge_expand}
 \vspace{1mm}
\end{figure}

Examples of such datasets are \textsc{gate},  \textsc{baffle}, \textsc{bugtrap} and \textsc{blob}. Here, \algName and \algHeuristic eliminate a large number of paths by only evaluating edges which are highly likely to be in collision and have several paths passing through them (\figref{fig:edge_expand}). In the 7D datasets, obstacles are highly concentrated near the goal region, which explains the strong performance of the uninformed \selectorBackward selector. However, due to a very large number of edges and limited training sets, \featPost and \featDeltaLength are inaccurate causing the learner to fail to outperform \selectorBackward.  

\begin{observation}
On datasets with uniformly spread obstacles, uninformed heuristics can perform better than \algName.
\end{observation}
Examples of such datasets are \textsc{twowall} and \textsc{forest} where the lack of structures makes features such as posterior uninformative. This combined with the non-realizability of the oracle makes it difficult for \algName to learn a strong policy.

\begin{figure}[!t]
\centering
    \includegraphics[width=\columnwidth]{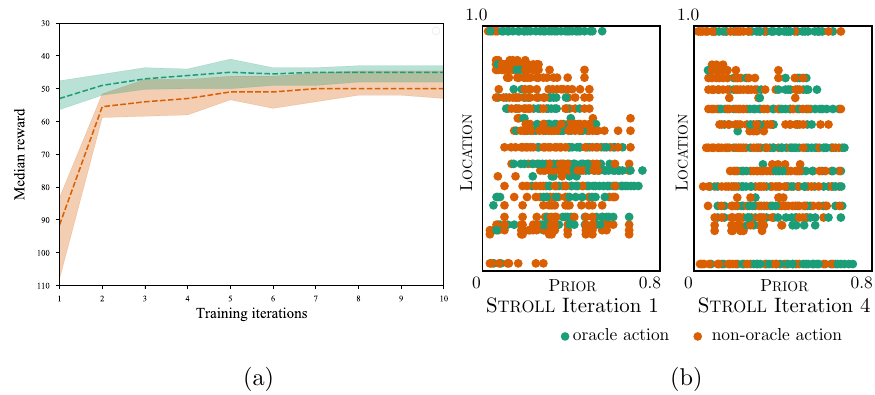}
\caption{ (a) \algHeuristic (green) vs \algName (orange) (b) Densification of data. }
    \label{fig:stroller_plots}
\end{figure}

\begin{figure}[!t]
    \centering
    \includegraphics[width=\columnwidth]{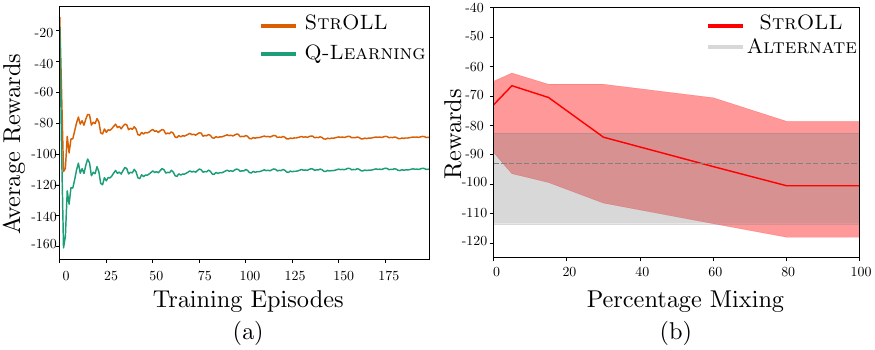}   
    \caption{(a) Running average reward for 200 episodes of training. Q learning suffers due to large state space and sparse rewards. (b) Performance on validation set of 200 worlds with contamination from different distribution. \vspace{-6mm}}
    \label{fig:q_learn_plot}
\end{figure}

{
	\renewcommand{\arraystretch}{1.5}
\begin{table*}[!t]
\small
\centering
\caption{ Edges evaluated by different algorithms across different datasets (median, upper and lower C.I on 200 held-out environments). Highlighted is the best performing selector in terms of median score not counting the oracle.}
\begin{tabulary}{\textwidth}{LCCCCCCCCC}\toprule
%& & \multicolumn{4}{c}{ {\bf Heuristic Baselines} }  &  \multicolumn{2}{c}{ {\bf Learning Baselines} } & {\bf Ours} \\
       & {\bf \selectorOracle}  & {\bf \selectorBackward}    & {\bf \selectorAlternate} & {\bf \selectorFailFast} & {\bf \selectorPostFailFast} & {\bf \selectorPDL}  &{\bf \supervisedAlg} & {\bf \algName} & {\bf \algHeuristic} \\ \midrule
\multicolumn{10}{c}{ {\bf 2D Geometric Planning} }   \\
\textsc{OneWall}    & $80.0^{+6.0}_{-48.0}$  & $87.0^{+8.4}_{-41}$   & $112.0^{+12.8}_{-60.0}$ &  $82.0^{+3.0}_{-47.0}$ & $81.0^{+3.0}_{-49.0} $ & $85.0^{+6.8}_{-52.6.0} $ & \cmark $79.0^{+3.0}_{-45.0}$ & \cmark $79.0^{+5.0}_{-44.8}$ &\cmark $79.0^{+5.0}_{-44.8}$  \\ 
\textsc{TwoWall}    & $107.0^{+23.0}_{-0.0} $ & $199.0^{+8.0}_{-19.0}$  & $138.0^{+7.0}_{-2.0}$   & $178.0^{+0.0}_{-6.0}$ & $177.0^{+0.0}_{-7.0}$ & \cmark $120.0^{+19.0}_{-0.0} $ & $177.0^{+0.0}_{-6.0}$ & $177.0^{+0.0}_{-6.0}$ &  $170.0^{+12.2}_{-0.0}$  \\ 
\textsc{Forest}    & $90^{+14.4}_{-10.0}$  & $128.0^{+15.0}_{-16.4}$   & $115.0^{+12.0}_{-13.2}$ & $135.0^{+13.0}_{-16.0}$ & $116.0^{+13.2}_{-16.4}$ & \cmark $102.0^{+15.0}_{-12.0} $& $117.0^{+19.2}_{-17.0}$ & $115.0^{+20.0}_{-15.0}$  & $115.0^{+21.2}_{-13.4}$  \\ 
\textsc{Gate}    & $50.0^{+6.0}_{-9.0}$ & $74.0^{+8.0}_{-9.0}$   & $75.0^{+14.2}_{-6.2}$ & $60.0^{+8.0}_{-6.2}$ & $50.0^{+7.0}_{-8.2}$ & $53.0^{+7.0}_{-9.2} $ & $50.0^{+7.0}_{-7.2}$ & \cmark$48.0^{+10.0}_{-7.2}$ & \cmark$48.0^{+9.2}_{-9.2}$  \\ 
\textsc{Maze}    & $537.0^{+37.0}_{-24.6}$ & $668.5^{+40.3}_{-56.1}$   &  $613.0^{+39.6}_{-33}$ & $512^{+52.2}_{-34.0}$ & $516.5^{+33.70}_{-36.50}$ & $529.0^{+40.0}_{-37.2} $ & \cmark $502.5^{+58.7}_{-28.2}$ & $512.0^{+42.0}_{-31.0}$ & $554.0^{+52.2}_{-59.0}$  \\ 
\textsc{Baffle}   & $219.0^{+18.0}_{-12.0}$ & $244.0^{+14.6}_{-6.0}$   & $311.0^{+8.0}_{-15.6}$ & $232.0^{+6.0}_{-12.0}$ & $211.0^{+7.8}_{-6.0}$ & $230.0^{+18.0}_{-17.0} $ &\cmark $206.0^{+6.8}_{-3.0}$ &\cmark $205.0^{+6.0}_{-3.0}$ &$207.0^{+7.0}_{-2.0} $  \\ 
\textsc{Bugtrap}  & $77.0^{+12.0}_{-9.4}$  & $104.0^{+6.0}_{-14.0}$  & $112.5^{+16.9}_{-11.5}$ & $90.5^{+10.9}_{-13.5}$ & $75.5^{+16.5}_{-6.5}$ & $84.5^{+12.9}_{-10.5} $ & \cmark $75.0^{+15.4}_{-6.4}$ & \cmark $75.0^{+15.4}_{-6.4}$ & \cmark $75.0^{+15.4}_{-6.4}$  \\ 
\textsc{Blob}  & $72.0^{+12.0}_{-4.0}$ & $92.0^{+5.4}_{-3.4}$   & $109.0^{+5.0}_{-7.0}$ &\cmark $70.0^{+6.0}_{-6.0}$ &\cmark $70.0^{+6.0}_{-6.0}$ & $80.0^{+9.0}_{-3.0} $ & $72.0^{+5.8}_{8.0}$ &\cmark $70.0^{+6.0}_{-6.0}$  &\cmark $70.0^{+6.0}_{-6.0}$  \\ 
\multicolumn{10}{c}{ {\bf 7D Manipulation Planning} }   \\
\textsc{Clutter1}  & $35.5^{+1.5}_{-1.5}$  & $38.0^{+12.0}_{-10.0}$   & $44.0^{+14.2}_{-4.0}$ & $92.0^{+5.0}_{-0.0}$ & $88.0^{+9.6}_{-0.0}$ & \cmark $37.5^{+2.1}_{-1.5} $ & $95.5^{+17.7}_{-11.1}$ & $50.0^{+3.0}_{-6.0}$ & $45.0^{+2.0}_{-1.6}$   \\ 
\textsc{Clutter2}  & $34.0^{+1.0}_{-2.0}$  & \cmark $32.0^{+3.0}_{-4.0}$   & $41.0^{+2.0}_{-2.2}$ & $85.0^{+0.0}_{-3.0}$ & $84.0^{+0.0}_{-2.0}$ & $37.0^{+0.0}_{-5.0} $ & $104.0^{+6.2}_{-6.8}$ & $47.0^{+2.0}_{-11.2}$ & $ 44.0^{+5.0}_{-7.2}$ \\ 
\bottomrule
\end{tabulary}
\label{tab:benchmark_results}
\end{table*}
}

\subsection{Case Studies}
\begin{ques}
How does performance vary with training data?
\end{ques}
\figref{fig:stroller_plots}(a) shows the improvement in median validation reward with an increasing number of training iterations. Also, \figref{fig:stroller_plots}(b) shows that with more iterations, the learner visits diverse parts of the state-space on x-axis not visited by the oracle.  

\begin{ques}
How significant is the impact of heuristic roll-in on stabilizing learning in high-dimensional problems?
\end{ques}
\figref{fig:stroller_plots} shows a comparison of the median validation return per iteration using \algName versus \algHeuristic on \textsc{clutter1} dataset. Heuristic roll-in helps converge to a better policy in lesser number of iterations. Interestingly, the policy learned in the first iteration of \algHeuristic is significantly better than \algName, demonstrating the stabilizing effects of heuristic roll-in. 

\begin{ques}
How does performance compare to reinforcement learning with function approximation?
\end{ques}
\figref{fig:q_learn_plot}(a) shows training curves for \algName and \textsc{Q-Learning} with linear function approximation and experience replay. \algName is more sample efficient and converges to a competitive policy faster.

\begin{ques}
How does performance vary with train-test mismatch?
\end{ques}
\figref{fig:q_learn_plot}(b) shows a stress-test of a policy learned on \textsc{one wall} by running it on a validation set which is increasingly contaminated by environments from \textsc{forest}. The learned policy performs better than the best uninformed heuristic on \textsc{forest} for up to $60\%$ contamination.
% \begin{figure}[!t]
% \centering 
% \includegraphics[width=0.5\linewidth]{figs/gen_plot.pdf}
% \caption{Generalization performance as a function of percentage of contamination of validation dataset from a different distribution. Solid red line is the median performance of the learned policy and dashed gray line is the median performance of the best heuristic on \textsc{forest} dataset, with shaded areas showing corresponding confidence intervals.}
% \label{fig:gen_plot}
% \end{figure}

% !TEX root = ../main.tex

\section{Related Work}
\label{sec:related_work}

In domains where edge evaluations are expensive and dominate planning time, a \emph{lazy approach} is often employed~\citep{bohlin2000path} wherein the graph is constructed \emph{without} testing if edges are collision-free. \lazysp~\citep{dellin2016unifying} extends the graph all the way to the goal, before evaluating edges. LWA*~\cite{cohen2015planning} extends the graph only a single step before evaluation. (LRA*)~\citep{Mandalika18} is able to trade-off between them by allowing the search to go to an arbitrary lookahead. The principle of laziness is reflected in similar techniques for randomized search~\citep{gammell2015batch,hauser2015lazy}. 

Several previous works investigated leveraging priors in search. FuzzyPRM~\citep{nielsen2000two} evaluates paths that minimize the probability of collision. The Anytime Edge Evaluation (AEE*) framework~\cite{narayanan2017heuristic} uses an anytime strategy for edge evaluation informed by priors. \bisect~\cite{choudhury2017active} and \direct~\cite{choudhury2018bayesian} casts search as Bayesian active learning to derive edge evaluation. However, these methods make specific assumptions about the graph or about the priors. Our approach is more general.

Efficient collision checking has its own history in the context of motion planning. Other approaches model belief over the configuration space to speed-up collision checking \citep{huh2016learning,choudhury2016pareto}, sample vertices in promising regions \citep{bialkowski2013free} or grow the search tree to explore the configuration space \citep{hsu1997path,burns2005sampling,lacevic2016burs}. However, these approaches make geometric assumptions and rely on domain knowledge. We work directly with graphs and are agnostic with respect to the domain.

Several recent works use imitation learning~\citep{ross2011reduction,ross2014reinforcement,sun2017deeply} to bootstrap reinforcement learning. THOR~\citep{sun2018truncated} performs a multi-step search to gain advantage over the reference policy. LOKI~\citep{cheng2018fast} switches from IL to RL. Imitation of clairvoyant oracles has been used in multiple domains like information gathering~\citep{choudhury2017data}, heuristic search~\citep{bhardwaj2017heuristic}, and MPC~\citep{kahn2016plato,tamar2016hindsight}.

% !TEX root = ../main.tex

\section{Discussion}
\label{sec:discussion}

% We have presented a general framework for lazy search (\glrastar). The framework is simple and interleaves two phases - search and evaluation. 
% In the search phase, it extends a lazy shortest-path tree forwards without evaluating any edges till an \Event is triggered. 
% It then switches to an evaluation phase. 
% It finds the shortest subpath to a leaf node of the tree and invokes a \Selector to evaluate an edge on it. 
% Judicious choice of \Event and \Selector allows one to balance search effort with edge evaluation to minimize overall planning time.

%We examined the problem of minimizing edge evaluations in lazy search on a distribution of worlds. To solve this problem, we first formulated the problem of deciding which edge to evaluate as a MDP.  Because the MDP is difficult to solve for large graphs, 
%we instead present an algorithm that learns policies by imitating clairvoyant oracles during training.
%This allows us to exploit a powerful structure to our problem - if the world is known, we can find such oracles that can optimally evaluate edges.   
%
We examined the problem of minimizing edge evaluations in lazy search on a distribution of worlds. We first formulated the problem of deciding which edge to evaluate as an MDP and
presented an algorithm to learn policies by imitating clairvoyant oracles, which, if the world is known, can optimally evaluate edges.
%This allows us to exploit a powerful structure to our problem - if the world is known, we can find such oracles that can optimally evaluate edges.   
While imitation learning of clairvoyant oracles is effective, the approach may be further improved through reinforcement learning~\citep{sun2018truncated,cheng2018fast}. There are two arguments for this. First, in practice we do not use the exact oracle but a sub-optimal approximation. Second, even if we could use the exact oracle, it may not be realizable by the policy. %On the other hand imitation learning provides a good initialization for policy improvement techniques~\citep{kakade2002approximately,levine2013guided}. 

\section*{Acknowledgments}

This work was (partially) funded by the National Institute of Health R01 (\#R01EB019335), National Science Foundation CPS (\#1544797), National Science Foundation NRI (\#1637748), National Science Foundation CAREER (\#1750483), the Office of Naval Research, the RCTA, Amazon, and Honda Research Institute USA.

\bibliographystyle{unsrtnat}
\bibliography{references}

\end{document}